\pdfoutput=1
\documentclass[letterpaper, 10 pt, conference]{ieeeconf}  

\IEEEoverridecommandlockouts                              

\usepackage{amsmath} 
\usepackage{amssymb}  
\usepackage{cite}
\usepackage{svg}
\svgpath{{../data/}}
\usepackage{url}
\usepackage{hyperref}
\usepackage{subcaption}
\usepackage{textcomp}
\usepackage{gensymb}
\usepackage{tabularx}
\usepackage{multirow}
\usepackage{threeparttable}
\usepackage{dsfont}
\usepackage{booktabs, caption, makecell}
\usepackage[free-standing-units=true]{siunitx}
\usepackage{amsmath}
\usepackage{tikz}
\DeclareMathOperator*{\argmax}{arg\,max}

\usepackage[most]{tcolorbox}
\usepackage{wrapfig}
\definecolor{ao(english)}{rgb}{0.0, 0.5, 0.0}
\definecolor{cocoabrown}{rgb}{0.82, 0.41, 0.12}
\usepackage{tikz}
\newcommand{\tikzcircle}[2][color=red!40,fill=red!40]{\tikz[baseline=-0.5ex]\draw[#1,radius=#2] (0,0) circle ;}%
\usepackage{changepage}

\title{\LARGE \bf
A Closer Look at Reward Decomposition for High-level Robotic Explanations 
}

\author{Wenhao Lu$^{1}$, Xufeng Zhao$^{1}$, Sven Magg$^{2}$, Martin Gromniak$^{1,3}$, Mengdi Li$^{1}$, Stefan Wermter$^{1}$
\thanks{This research was funded by the Federal Ministry for Economic Affairs and Climate Action (BMWK) under the Federal Aviation Research Programme (LuFO), Projekt VeriKAS (20X1905)}
\thanks{$^{1}$The authors are with Knowledge Technology Group, Department of Informatics, University of Hamburg}
\thanks{$^{2}$The author is with Hamburger Informatik Technologie-Center e.V. (HITeC)
        }%
\thanks{$^{3}$The author is with ZAL Center of Applied Aeronautical Research
        }%
 }

\begin{document}

\maketitle
\thispagestyle{empty}
\pagestyle{empty}

\begin{abstract}
Explaining the behaviour of intelligent agents learned by reinforcement learning (RL) to humans is challenging yet crucial due to their incomprehensible proprioceptive states, variational intermediate goals, and resultant unpredictability.
Moreover, one-step explanations for RL agents can be ambiguous as they fail to account for the agent's future behaviour at each transition, adding to the complexity of explaining robot actions.
By leveraging abstracted actions that map to task-specific primitives, we avoid explanations on the movement level. To further improve the transparency and explainability of robotic systems, we propose an explainable Q-Map learning framework that combines reward decomposition (RD) with abstracted action spaces, allowing for non-ambiguous and high-level explanations based on object properties in the task.
We demonstrate the effectiveness of our framework through quantitative and qualitative analysis of two robotic scenarios, showcasing visual and textual explanations, from output artefacts of RD explanations, that are easy for humans to comprehend.
Additionally, we demonstrate the versatility of integrating these artefacts with large language models (LLMs) for reasoning and interactive querying.

\end{abstract}

\section{Introduction}


The developmental progress of artificial intelligence highlights the crucial role of transparency as a fundamental component in its advancement \cite{Arrieta20ExplainableArtificial,Heuillet21ExplainabilityDeep,Qing22SurveyExplainable, SLLKW23, KABLSSGWW22}.
Providing explanations for deep learning models used in robotic tasks such as navigation \cite{gu2022multi} and manipulation \cite{DBLP:journals/corr/abs-2110-03655} is crucial for establishing trust and understanding between humans and robots. Despite significant advancements in enhancing end-to-end models and in blooming explanation techniques \cite{DBLP:journals/corr/SelvarajuDVCPB16, Juozapaitis2019ExplainableRL, kim2018interpretability}, it remains unclear and underexplored how to extract effective explanations of the robot's learned behaviours that are comprehensible to humans.

\begin{figure}[thpb]
      \centering
      \includegraphics[scale=0.2]{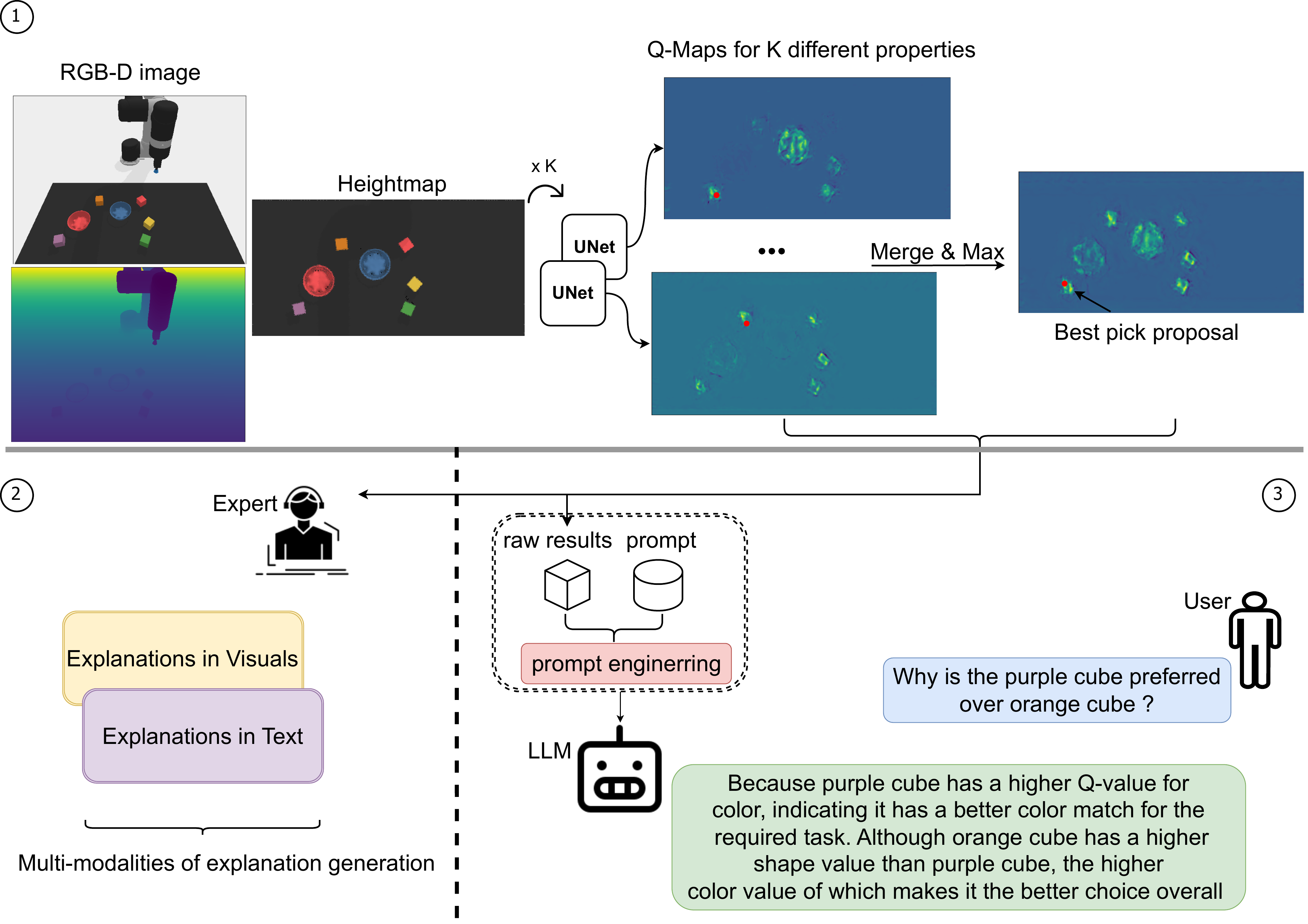}
      \caption{\textbf{Overview of our explainable Q-Map learning framework} for the grasping task of objects with a specific shape, in a predefined order of colours (e.g., picking up a purple cube before an orange one).
      Part 1 shows the learning of $K=2$ Q-Maps using FCN \cite{DBLP:journals/corr/LongSD14} to pick objects, with actions being primitives executed at pixels. Actions are evaluated based on $K$ component values, and the selection is explained by the contribution of each component. Interpreting these raw results can be handled by either human experts (part 2) or a large language model (LLM) (part 3). Human experts can interpret Q-Maps by visualizing (see Fig. \ref{fig:x}) or using text templates (Sec. \ref{sec: visual properties}). Alternatively, prompt engineering (see Table \ref{tab:prompt}) for an LLM provides interactive language explanations to its users. 
      }  
      \label{fig:learning framework in ravens}
\end{figure}

The level of understandability in an explainable system relies on more than just the transparency of the model. It also depends on a varying \textit{audience} with different cognitive skills and goals \cite{Arrieta20ExplainableArtificial,Heuillet21ExplainabilityDeep}. Similar to hierarchical RL \cite{DBLP:journals/corr/BaconHP16}, where a problem can be decomposed into various abstract levels of sub-problems, determining the appropriate level of behavioural abstraction is a crucial aspect of interpretability. Given that robotic tasks involve agents operating in a high-dimensional, low-level action space, such as positional movements or joint torques, with a visual representation of the environment, the agent's low-level actions are the natural subject of explanation.
However, one-step explanations on this level might be ambiguous, as demonstrated by previous research \cite{cite-key}, as the agent's future goals could be implicit at each transition. For example, in a grasping task with multiple objects to the left of the robot's gripper, the intention behind an action, e.g., \textit{move left}, may not be immediately discernible. This lack of clarity arises as the execution of that action can potentially lead to multiple (sub-)goals being pursued subsequently, creating ambiguity when explaining the robot's decision. We seek to elucidate the robotic system by focusing on the high-level space, which aligns with the slow and conscious thinking processes of humans. In contrast, the low-level action space, associated with the fast and unconscious thinking system, poses challenges due to its ambiguity and is inadequate for explanatory purposes~\cite{kahneman2011thinking,Gregor17VariationalIntrinsic}.
Hence, we propose to interpret the decision-making (robot) policy in abstracted (high-level) action spaces that correspond to task-specific motion primitives \cite{DBLP:journals/corr/abs-2010-01083} executed at a 3D position, referred to as an unambiguous (sub-)goal. To illustrate, the outcome of executing a grasping primitive (i.e., action) could be either a red cube or an empty result, as demonstrated in the previous example.
An abstracted action space comes with advantages, including a more specific goal context of humans' interests, decreased sampling complexity during the learning process, and enhanced generalizability due to the concise representation of the task and the ability to reuse primitives \cite{truong2022rethinking}.


Taking a kitchen scenario as an example, in which a robot is given an instruction to \textit{``find a container for a piece of cake''}, humans may be curious about why the robot selects a plate over a bowl, but may not be interested in the precise sequence of movements involved in picking up the plate. Additionally, humans require explanations that can be understood through cues readily available to them. The internal state of a robot, such as joint configurations and dynamics, is usually not accessible to humans. An explanation at this low-level space is hard for humans to comprehend, thus being ruled out of consideration.

A typical reinforcement learning (RL) process includes an agent continually pursuing rewards from the environment and updating the policy accordingly to maximize the expected future reward.
Thus, a closer look at the reward function is significantly helpful for understanding the inherent RL process.
However, rewards emerge from diverse aspects but are often treated as a scalar for RL.
Among the many previous works that try to establish explainable RL frameworks, \cite{
10.5555/3041838.3041921, DBLP:journals/corr/abs-1711-00138, DBLP:journals/corr/abs-1809-05630, DBLP:journals/corr/abs-1809-06061, DBLP:journals/corr/abs-1912-12191, Juozapaitis2019ExplainableRL, HEUILLET2021106685,Qing22SurveyExplainable}, the reward decomposition (RD) methods \cite{10.5555/3041838.3041921, Juozapaitis2019ExplainableRL, lin2020rd} help facilitate a more thorough understanding of the RL process by decomposing and learning from the different aspects of the environment. These works demonstrate the benefits of RD with toy examples and Atari games. However, when it comes to more complex scenarios such as robotic control, the challenges faced will be distinct in nature (e.g., high-dimensional visual perception). 
Particularly, object properties \cite{DBLP:journals/corr/abs-2010-14406, DBLP:journals/corr/CalliWSSAD15}, e.g. material, shape, weight, colour, etc., all contribute to making a decision by the robot. In addition, the inclusion of a flexible interface, such as a visual presentation or a querying dialogue, is beneficial for humans to effectively interact with the system.

As a whole, we construct an inherent \textit{explainable learning framework} (Fig.~\ref{fig:learning framework in ravens}, Sec.~\ref{explainable Q-Map learninig framework}) that 1) incorporates an RD process, 2) offers a corresponding multimedia interface, and 3) enables unambiguous and flexible RD-based explanation for robotic scenarios (Sec.~\ref{sec:exp}). 
Within this framework, the explanatory artefacts resulting from the reward decomposition are converted into interpretable visuals and templated languages (Sec. \ref{sec: visual properties}). These representations can be readily comprehended by humans or digested by Large Language Models (LLMs) to facilitate automatic reasoning and interactive querying (see Fig.~\ref{fig:x} and Fig.~\ref{fig:x drone} for example pipelines).
The effectiveness of our framework, in terms of task completion and explanation generation, is demonstrated via two distinct robotic scenarios: a grasping task performed by a robot arm and a landing task requiring a flying agent to search for suitable sites. 


\section{Related Work}

\textbf{Explainable RL (XRL).}
A commonly accepted classification for XRL methods is based on the distinction between post-hoc explainability and transparent models \cite{HEUILLET2021106685,Arrieta20ExplainableArtificial,Qing22SurveyExplainable, Juozapaitis2019ExplainableRL}.
Saliency map, a post-hoc explanation model, utilizes the gradient \cite{DBLP:journals/corr/SelvarajuDVCPB16} of actions with respect to the state to identify important features, which can be visualized as the saliency of pixels to a particular action. However, this method neither accounts for the agent's future intention nor does it truly reflect its behaviour \cite{DBLP:journals/corr/abs-1912-05743}. 
To establish a transparent XRL model, there are works that either construct surrogate models \cite{HEUILLET2021106685} or decompose received rewards \cite{Juozapaitis2019ExplainableRL}. By analyzing the various aspects considered within the reward function, this inherent ambiguity can be possibly eliminated.
The concept of reward decomposition was first proposed by \cite{10.5555/3041838.3041921}, and decomposes a complex task into simpler sub-tasks and assigns sub-rewards to them. \cite{Juozapaitis2019ExplainableRL} further extends this approach to explain the agent's behaviour with additional explanation metrics based on sub-rewards.


\textbf{Task Planning followed by Motion Planning.}
Abstracted action space is commonly used in task reasoning along with available low-level skills for motion planning \cite{DBLP:journals/corr/abs-2010-01083}. To enable a smooth integration of functional high-level modules like LLM and visuomotor controllers, prior work \cite{truong2022rethinking, ahn2022can} relies on pre-trained low-level skills in their learning pipeline. Our work, on the other hand, tackles the ambiguity issue inherent to one-step RL explanations by utilizing abstracted actions.





\textbf{Grounding LLMs.}
Large Language Models are showing impressive performances in many fields such as comprehending documents \cite{Brown20LanguageModels,Bubeck23SparksArtificial}, commonsense reasoning \cite{DBLP:journals/corr/abs-2201-11903}, and robotic planning \cite{ahn2022can,Zhao23ChatEnvironment}.
In order to further extend the post-hoc interpretation of the transparency of reward decomposition, we give an LLM access to the explanation artefacts of learned models, i.e. the templated language deemed for proficient users, resulting in an enhancement of usability and flexibility of interactive explanation querying.



\section{Preliminaries}
\subsection{Markov Decision Process}
We assume a Markov Decision Process (MDP) \cite{puterman2014markov} for the RL agent's interaction with the environment, including the state space $\mathcal{S}$, action space $\mathcal{A}$, state transition probabilities $\mathcal{P}$, action-state dependent reward function $\mathcal{R}$ and discount factor $\gamma$, written as $(\mathcal{S}, \mathcal{A}, \mathcal{R}, \mathcal{P}, \gamma)$. The objective of the MDP is to learn a policy $\pi: \mathcal{S}  \rightarrow  \mathcal{A}$ mapping from state to action that leads to a maximization of the expected cumulative reward $\mathbb{E}[\sum_{t=0}^{T} \gamma^t r_t]$. 

\subsection{Deep Q-learning Network}
Among others, one generic way to learn $\pi$ for an MDP is to first learn an action-value function $Q(s, a)$ \cite{Watkins:89}, which represents the expectation of a discounted return when the agent takes action $a_t$ in state $s_t$ and then follows $\pi$ in the future. Formally, it is written as 
\begin{equation*}
    Q(s_t, a_t) = \mathbb{E}_{\pi}[\, r_t + \gamma \max_{a_{t+1}} 
    Q(s_{t+1}, a_{t+1}) ]\, ,
\end{equation*}
and a greedy policy can be derived by taking the maximum over the action-value function $\pi = \argmax_{a_t} Q(s_t, a_t)$. Building on deep Q learning \cite{mnih2015humanlevel}, we approximate the value function $Q_\phi$ with a neural function approximator parameterized by $\phi$. The parameters $\phi$ are optimized by minimizing the loss
\begin{equation*}
\begin{split}
    J(\phi) & = \mathbb{E}_{(s_t, a_t, r_t, s_{t+1}) \sim \mathcal{D}}[(r_t \, + \\
    & \gamma Q_{\phi^\prime} (s_{t+1}, \argmax_{a_{t+1}}Q_\phi(s_{t+1}, a_{t+1})) - Q_\phi(s_t, a_t))^2],
\end{split}
\end{equation*}
in batch-mode (by sampling batch-sized transition tuples from a replay buffer $\mathcal{D}$) where $Q_{\phi^\prime}$ is a target network and will be periodically updated by copying the network $Q_\phi$.

\section{Problem Formulation} 

Most robotic tasks are modelled as an MDP with action spaces being continuous and reward functions being binary and non-decomposable, indicating success or failure of the task. 
Nevertheless, in real-world robotic tasks, rewards can originate from various aspects or sources.
We demonstrate environments, in which the robot operates, populated with a variety of objects of diverse properties (or concepts) like shape and visual appearance (e.g., colour and texture). Those semantic (high-level) concepts serve as key references to differentiate one object from others and thus can be used as sources of reward. The now decomposable rewards, along with the abstract action spaces, underlie our task learning framework which allows the (high-level) decision-making to be better explained, as no low-level motions are involved. Further, we deploy the RD technique to complement this framework with task-contextualized explanations. 

Notably, the modification implicitly renders explanations in (continuous) robot domains possible, by virtue of the RD \cite{Juozapaitis2019ExplainableRL}. We provide a formal description of how the abstract action spaces and the RD fit into an explainable learning framework in the next sections. 

\section{Explainable Q-Map Learning Framework} \label{explainable Q-Map learninig framework}

Our framework is named \textbf{eXplainable Q-Map Learning} (X-QMap\footnote{https://x-qmap.github.io/}), where properties pertaining to sub-rewards in the RD approach are utilised to guide the learning of \textit{Q-Maps }(2D matrices of Q-values) and provide the rationale for the decision-making. 
A collection of learned Q-Maps, with each relating to a specific object property, enables explanations at the property level and reveals the subset of features that contribute to a decision.

\subsection{Decomposed Rewards and Primitive Actions} \label{sec:decomposed rewards}


\textbf{Decomposable Rewards.}
We assume a composite reward function, composed of independent sub-rewards (components), and each of which accounts for a specific semantic property that influences a decision-making procedure.
As an example, criteria for a flying agent to land in appropriate locations could consist of 1) \textit{being on flat surfaces} and 2) \textit{being away from crowds}. Each criterion can be represented as a sub-reward $r_k$ which serves as an \textit{indicator} function
\begin{equation*}
\mathbb{I}(C_k) =
\begin{cases}
1 & \text{if } C_k \text{ is satisfied} \\
0 & \text{otherwise}, 
\end{cases}
\end{equation*}
denoting whether a task-specific property $C_k$  was identified~\footnote{For non-binary properties, a continuous reward $r \in [\,0, 1\,]$ can be used to indicate the rate of fulfillment.}.
The overall task reward is obtained by summing up all sub-rewards $\mathcal{R} = \sum_{k=0}^K r_k$.


\textbf{Primitive Actions.} Following \cite{DBLP:journals/corr/abs-1803-09956}, each robot action $a$ is parameterized as a motion primitive $\zeta$ that is executed at a 3D position projected by a pixel $P$ of an orthographic RGB-D heightmap of the scene, i.e., $a = ( \,  \zeta, \, P ) \,$.
The decoupling of the high-level semantic hierarchy from the low-level dynamics through this setting of primitive actions enables a focused analysis of properties and rewards that hold greater significance for users.

Our framework would benefit from a systematic routine of how to decompose task rewards, by extending approaches like \cite{lin2020rd, kim2018interpretability}, which extracts reward channels from agent interaction data, and we defer this to future work as the focus of this work lies in exhibiting a complete framework for robotic explanations.



\subsection{Learning Q-functions with Reward Decomposition} \label{reward decomposition}




The usage of off-policy reinforcement learning, e.g., Q-learning as explored in RD explanation \cite{Juozapaitis2019ExplainableRL}, allows for explaining the agent's preference for specific actions.
We, therefore, train an RL agent to learn a Q-function for evaluating areas of interest in the scene at any time step $t$, which takes the RGB-D as inputs, denoted by $s_t$. The Q-function takes the form of a dense pixel-wise 2D map of Q-values, known as \textit{Q-Map}, with the same dimensions as the input. This idea has been explored in works \cite{DBLP:journals/corr/abs-1710-01330, DBLP:journals/corr/abs-1803-09956, DBLP:journals/corr/abs-2105-14829}.

To handle multiple sub-rewards, we use a set of $K \in \mathbb{N}$ Q-Maps,
each supervised by a single reward component $r_k$. The optimal (global) action $a^\ast$ at a state $s_t$ is the pixel with the highest value over the summed Q-functions across all $K$ component values $Q_{\phi^k}$, i.e.,
\begin{equation}
\label{eq:astar}
    a^\ast = \argmax_{a_t^{\prime}} \sum_{k=1}^K Q_{\phi^{k}}(s_t, a_t^{\prime}).
\end{equation}
Each action value function $Q_{\phi^{k}}$ is approximated by a separate fully convolutional network (FCN) \cite{DBLP:journals/corr/LongSD14} $\phi^k$, based on an UNet architecture \cite{DBLP:journals/corr/RonnebergerFB15}.
As per findings by \cite{10.5555/3041838.3041921  }, individually trained Q-functions on different reward components can result in behaviour (by summing up these component values) that is equivalent to training on the sum of all components. 
This equivalence is assured through bootstrapping the global action for the next state $s_{t+1}$ when computing TD-error $\delta^k_t$ for updating each Q-Map.

\begin{equation*}
\delta^k_t = r_k(s_t, a_t) + \gamma Q_{\phi^{k}}(s_{t+1}, a^\ast) - Q_{\phi^{k}}(s_t, a_t)
\end{equation*}



\subsection{Visual Properties: Key to Explanations} \label{sec: visual properties}

We use the learned $K$ Q-Maps to construct two types of explanations 
for global action selection. The first \textit{shallow} explanation refers to \textit{why a specific action was chosen}, while a progressive \textit{contrastive} explanation explains \textit{why one action was preferred over another}. We compare the global action to other possible candidates by analyzing the highest-scoring pixels in each individual component Q-Map, enabling action comparisons.

Analyzing the values of each component Q-Map for the global action $a^\ast$ can provide insight into the quantitative contribution of each component $Q_{\phi^{k}}(s, a^\ast)$ to the decision.
The component with the highest Q-value among all components is the one that contributes the most: $k = \argmax_k Q_{\phi^{k}}(s, a^\ast)$ in a given state, providing a simple answer as the shallow explanation. 
When contrasting one action $a_i$ with another $a_j$, we adopt reward difference explanation (RDX) as in \cite{Juozapaitis2019ExplainableRL} to gain insight into the Q-value difference between two actions under each component: $\Delta_k(s, a_i, a_j) = Q_{\phi^{k}}(s, a_i) - Q_{\phi^{k}}(s, a_j)$. RDX quantitatively represents the advantage or disadvantage of $a_i$ over $a_j$ under each component, thus forming answers as the contrastive explanation. 

To ensure the explainability of our system for a broad range of \textit{audience} \cite{Heuillet21ExplainabilityDeep}, the explanations are presented through both visual charts and language descriptions.

\textbf{Explanations in Visual Charts.} 
We use a bar chart to visualize the statistics corresponding to $K+1$ actions of interest, including the overall optimal, \textit{Selected} action $a^{*}$ of the composite Q-Map (see Eq.~\ref{eq:astar}) and sub-optimal actions $a_k^{*} = \argmax_{a} Q_{\phi^k}(s_t, a)$ of a biased consideration on the component $C_k$.

The height distribution of bars in the plot indicates the importance of each component and constitutes a shallow explanation. Examples are depicted in Fig.~\ref{fig:x} and Fig.~\ref{fig:x drone} where $K=2$.
Analogously, we visualize RDX over pairs: \textit{Selected} vs. $A$, \textit{Selected} vs. $B$, and $A$ vs. $B$ as another bar plot. This plot could visually demonstrate Q-value differences among actions under different components, thus clearly forming a visual explanation of the contrastive question 

\textbf{Explanations in Language Claims.}
Though visual explanations in bar plots provide intuitive representations of Q-values over selective actions, their accessibility (e.g., a broader range of \textit{audience} \cite{Heuillet21ExplainabilityDeep}), usability (e.g., integration with other services), and flexibility (e.g., querying with a specific purpose) can be further enhanced by incorporating language. This is due to that the universal representation ability of language\cite{Bubeck23SparksArtificial} enables a more inclusive and adaptable potential.
Therefore, we propose to use templated language to accompany visuals with plain text that clarifies the contribution of object properties to each action selection.

Let $\chi$ be a set of properties inferred in the scene, $\{A, B, C, \dots \}$ be the action (pixel) choices that feature the highest values in individual Q-Maps, and $\{ O_A, O_B, O_C, \dots \}$ be the corresponding objects. A possible language-based reasoning template for explaining why choosing pixel $C$ (or object $O_C$) is $\Rightarrow$: \textit{\{object $O_C$\} owns the highest Q-value in the current scene, with \{feature $\chi$\} component contributing most to the selection}. This template serves shallow explanations and the one set up for contrastive explanations is $\Rightarrow$: \textit{In contrast to \{action $A$\} (\{object $O_A$\}), \{object $O_C$\} is chosen due to its \{feature $\chi_1$\}, not due to its \{feature $\chi_2, \dots$\} or \{object $O_C$\} is chosen due to its \{feature $\chi$\}}. 
The completion of placeholders above is based on elaboration on bar plots by programming.

\textbf{LLM-enabled Language Explanation.} The usage of templates makes it easier to communicate explanations to a wider group of practitioners. However, recipients cannot query templates to resolve potential confusion through interaction. We argue that explanations should be diverse in the sense that every recipient can receive customized explanations as the focus varies from person to person. LLMs have been showing impressive ability in tasks requiring contextual reasoning~\cite{ahn2022can}. Nevertheless, the exploration of their potential within the context of XAI still requires increasing attention. 

\begin{table}[ht!]
\centering
\caption{A case example with fixed prompt (\tikzcircle[color=gray!55, fill=gray!55]{2.5pt}), human query (\tikzcircle[color=black!75, fill=black!75]{2.5pt}), templated artifacts of RD explanation (\tikzcircle[color=cocoabrown!55, fill=cocoabrown!55]{2.5pt}) and LLM responses (\tikzcircle[color=ao(english)!55, fill=ao(english)!55]{2.5pt}).
}
\begin{adjustwidth}{-0.2cm}{0.2cm}
\begin{tabularx}{\linewidth}{l}
\begin{tcolorbox}[
    fonttitle=\fontsize{9}{12}\selectfont, 
    colbacktitle=gray,
    colframe=gray,
    boxrule=0.5pt,
    standard jigsaw,
    opacityback=0,  
    frame hidden,
    interior hidden,
    boxsep=0pt,
    left=4pt,
    right=4pt,
    top=4pt,
    bottom=4pt,
    fontupper=\linespread{0.8}\selectfont,
    fontlower=\linespread{0.8}\selectfont,
]
{\small
\textbf{System:}
\textcolor{gray}{Context: Imagine there is a visual pick-up task that a robotic arm needs to learn to solve. The objective of the task is to pick up objects with task-specific properties. We train an agent to achieve this using Q-learning which outputs a 2D matrix of Q-values of the same size as the input image. The Q-values quantitatively describe the utility of action (pixel) choices, each corresponding to executing the pick-up primitive at a 3D position mapped from that pixel. The Q-value of every action (and its associated object) is further decomposed into two component values, one evaluating its score in being a cube (not a bowl) and the other being in which color ranking (red $<$ orange $<$ yellow $<$ green $<$ blue $<$ purple), summing up to its overall Q-value.\\
\\
You are helping humans understand the action choices of the trained Q-agent given a scene of the task. In each turn, you are provided with three action choices along with their component values and overall values of the scene.\\
The user will ask you two types of questions:\\
1) shallow question - why is an action chosen?\\
2) contrastive question - why is one action preferred over another?\\
Please answer those questions by text and keep the text simple and clear.\\
\\
Scene Description:} \textcolor{cocoabrown}{Three pixels A, B, Selected are given, where A = a blue cube, its values = \{color: 0.577, shape: 0.426, overall: 1.003\}, B = a red cube, its values = \{color: 0.017, shape: 0.745, overall: 0.762\}, Selected = a blue cube, its value = \{color: 0.557, shape: 0.516, overall: 1.073\}. The value difference RDX for action pairs in each component: (Selected, A) = \{color: -0.02, shape: 0.09\}, (Selected, B) = \{color: 0.54, shape: -0.229\}.}\\
\textbf{Human:} \textcolor{black}{Now pixel Selected is chosen, and the shallow question is: why is pixel Selected chosen to pick up?}\\
\textbf{AI:} \textcolor{ao(english)}{The pixel Selected is chosen to pick up because it has the highest Q-value overall among the three pixels, indicating it is the best choice based on its shape and color values.}\\
\textbf{Human:} \textcolor{black}{Contrastive question: why is pixel Selected preferred over pixel B?}\\
\textbf{AI:} \textcolor{ao(english)}{Pixel Selected is preferred over pixel B because it has a higher Q-value for color, indicating it has a better color match for the required task. Although pixel B has a higher shape value than Selected, the higher color value of Selected makes it the better choice overall.}
}
\end{tcolorbox}
\label{tab:prompt}
\end{tabularx}
\end{adjustwidth}
\end{table}
We thus emphasize the use of LLMs for generating task-specific language explanations to potential users. 
To achieve this, we provide an LLM with prompts that include a task summary and short examples for better understanding.
Specifically, we describe in words what is expressed by bar plots of Q-values for a specific scene and pose questions to the LLM to initiate the conversation. ChatGPT \cite{chatgpt} is used to implement this dialogue, aiming to enable an interactive understanding of robot behaviour for diverse and flexible applications. See Table~\ref{tab:prompt} for an example with a detailed prompt. 

\section{Experiments}
\label{sec:exp}
We tested our proposed explainable framework in two distinct robotic scenarios: one with a robot arm deducing
the next objects to grasp; and the other with a flying agent searching for optimal landing sites. 
The experiments are conducted to answer the following questions:
\begin{enumerate}
    \item Does the combination of abstract action space and property-linked sub-rewards allow for successful task training in robotics?
    \item How can the learned decomposed Q-Maps be exploited to generate explanations for the agent's action choice?
\end{enumerate}

\subsection{Simulation Setups}


In accordance with \cite{DBLP:journals/corr/abs-2010-14406}, a heightmap image is used as the state representation $s_t$ 
in both scenarios. This heightmap is retrieved by unprojecting RGB-D data to a 3D point cloud, followed by its transformation into an orthographic projection. Each pixel $(u, v)$ corresponds to a predefined 3D spatial window in this projection. As aforementioned, we abstract the action space by parameterizing each action with a motion primitive and the 3D position at which it is executed. Both scenarios have a built-in primitive for landing and pick-up behaviour, respectively. The properties used are described in Table \ref{table:sub-rewards collection}.

\begin{table}[th]
\centering
\caption{Property description along different dimensions used in the two scenarios.}

\begin{tabular}{c| c c c}
\toprule
\thead{Object} & \thead{Appearance} & \thead{Geometry} & \thead{Texture}\\[0.5ex] 
\midrule
\midrule
Cube  &  rainbow & square & -\\ 
Bowl  & rainbow  & rounded & -\\ 
Block  & green, red, blue, gray & - & flat or non-flat\\ 

\bottomrule
\end{tabular}

\label{table:sub-rewards collection}
\end{table}



\begin{figure}[thpb]
      \centering
      \includegraphics[scale=0.13]{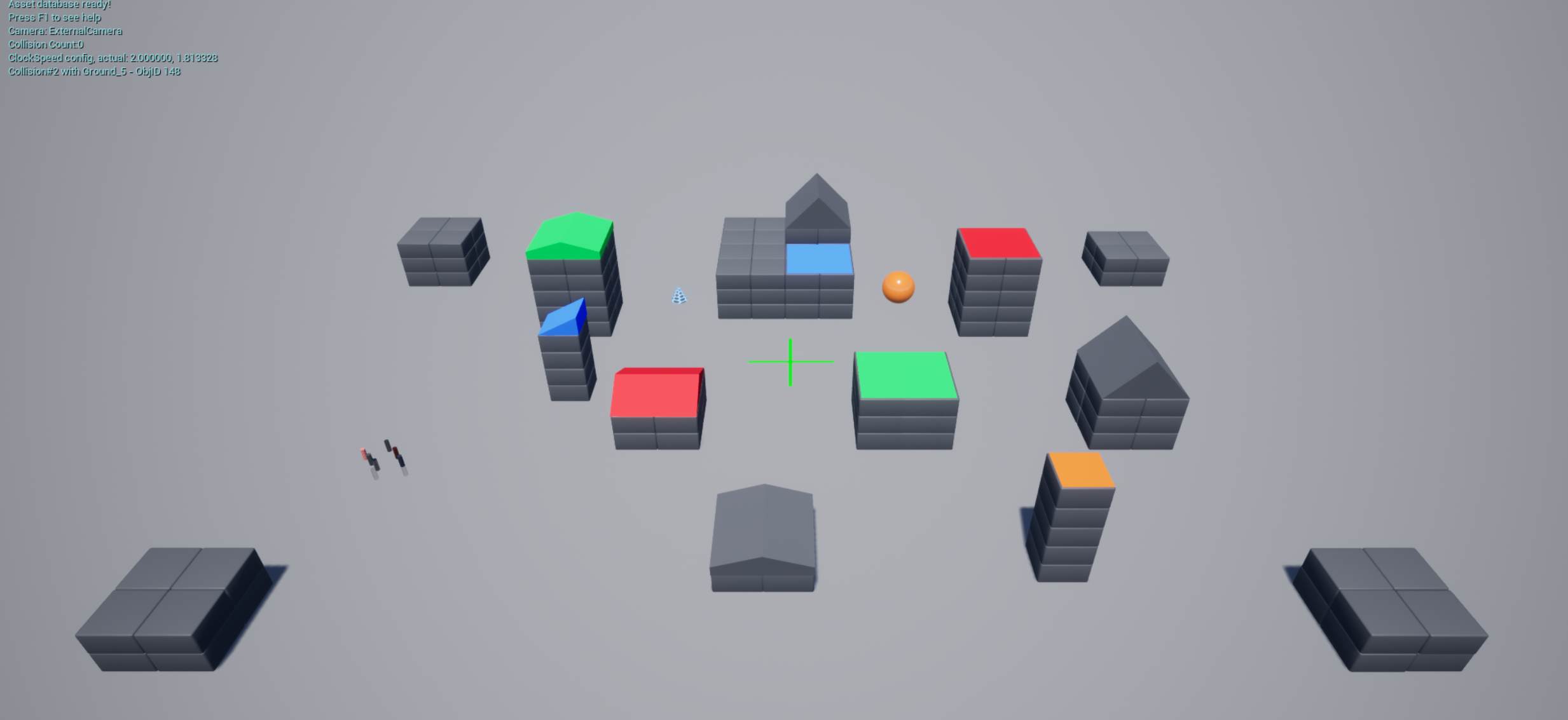}
      \caption{Bird's-eye view of Airsim's Blocks environment. Blocks can have sides inclined and coloured to indicate diverse consequences when landing.
      }
      \label{fig:blocks environment}
\end{figure}

\textbf{The Landing Scenario.}
In the Blocks environment (Fig.~\ref{fig:blocks environment}) of Airsim \cite{DBLP:journals/corr/ShahDLK17}, a custom flying agent is tasked with finding suitable landing spots that satisfy human-specified criteria. As a demonstration of the applicability of the Q-Map framework, we focus on two properties for landing spots: surface flatness and colour (excluding grey). We choose the two to have easily interpretable features for humans. 

The observation is rendered as a heightmap of resolution $84 \times 84$, showing a partial view of the Blocks scene. The reward function comprises two binary sub-reward functions, one assigns a reward of 1 if the 3D spot projected by the pixel is flat, and the other assigns another 1 if the spot is also coloured (not in grey). We use the formula $\theta = \arccos(\vec{n} \cdot \vec{q})$ to measure the flatness of a surface at a landing spot, where $\vec{n}$ is the vertical unit vector of the plane and $\vec{q}$ is the surface normal (of length 1) facing upwards at that point. We consider $\theta\leq \ang{5}$ as flat.
Each episode involves the flying agent selecting a landing spot (pixel) and executing a touchdown within 50 steps before being reset to a new 3D location with a constant altitude.

\begin{figure}[!htb]
    \centering
    \includegraphics[scale=0.4]{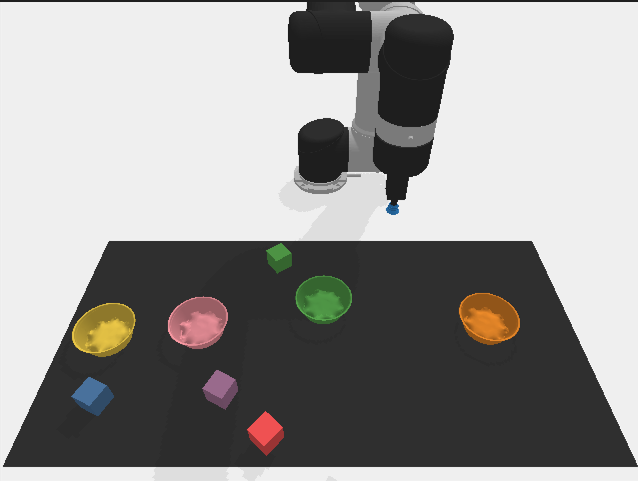}
    
     \caption{Example grasping tasks in Ravens. Objects including colored cubes and bowls are cluttered on the tabletop with a UR5e robot nearby.}
    \label{fig:ravens tasks}
\end{figure}

\textbf{The Grasping Scenario.} In this scenario, a simulated Universal Robot UR5e with a suction cup gripper is used for a pick-and-place task from Ravens benchmark \cite{DBLP:journals/corr/abs-2010-14406} (depicted in Fig.~\ref{fig:ravens tasks}). The task goal is to find pickable areas for task-specific objects and use the suction cup to grasp them. The task training involves 7 or more objects with randomly assigned shapes and colours (forming a rainbow spectrum) in each episode. Bad initialization (with unreachable targets under the arm's joint configurations) is filtered out at first, and the episode ends when the workspace is void of desired objects (i.e., coloured cubes).

The three RGB-D cameras pointing to the tabletop are rendered into a $320 \times 160$ resolution heightmap. Sub-rewards based on the object's \textbf{colour} and \textbf{shape} are constructed. To avoid handcrafting heuristics, we incorporate a simple rule into the learning process: suctioning a bowl results in a failure with a sub-reward shape of 0; whereas grasping a square results in a sub-reward shape of 1. The object's colour is rated on a scale between 0 and 1 based on the ranking of the object's colour in a rainbow spectrum (e.g., $\frac{1}{5}$ for orange) when any object is grabbed.

\begin{figure*}[!htb]
    \centering
    \includegraphics[scale=0.17]{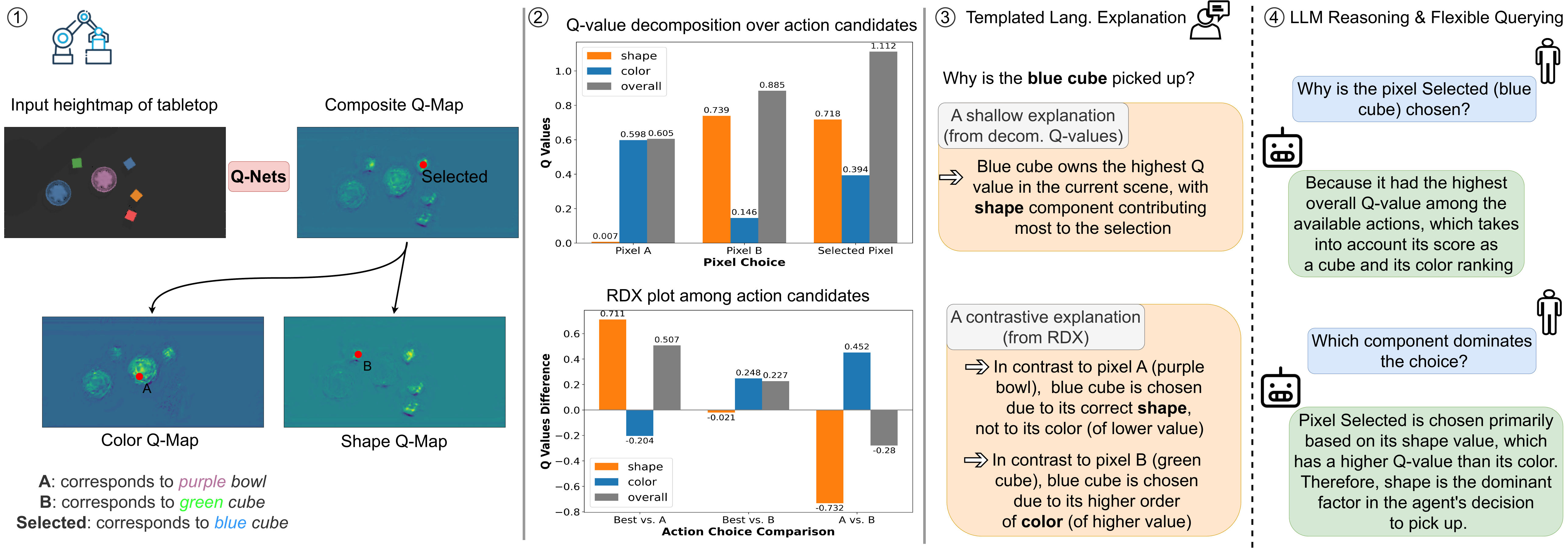}
     \caption{Illustration of the learned Q-Maps for the grasping scenario and the multiple modalities of explanations. Part 1 shows the raw data comprising Q-Maps output to the task scene. Part 2 provides a visual summary of Q-values for pixel choices A, B, and Selected, which visually inform shallow and contrastive explanations. These are further converted into language explanations using templates (part 3), serving as a supplement to the understanding of the agent's behaviour. Additionally, part 4 showcases how LLM facilitates users' understanding of agents' behaviour via question-answering interactions.
     }
    \label{fig:x}
\end{figure*}
\begin{figure*}[!htb]
    \centering
 \includegraphics[scale=0.16]{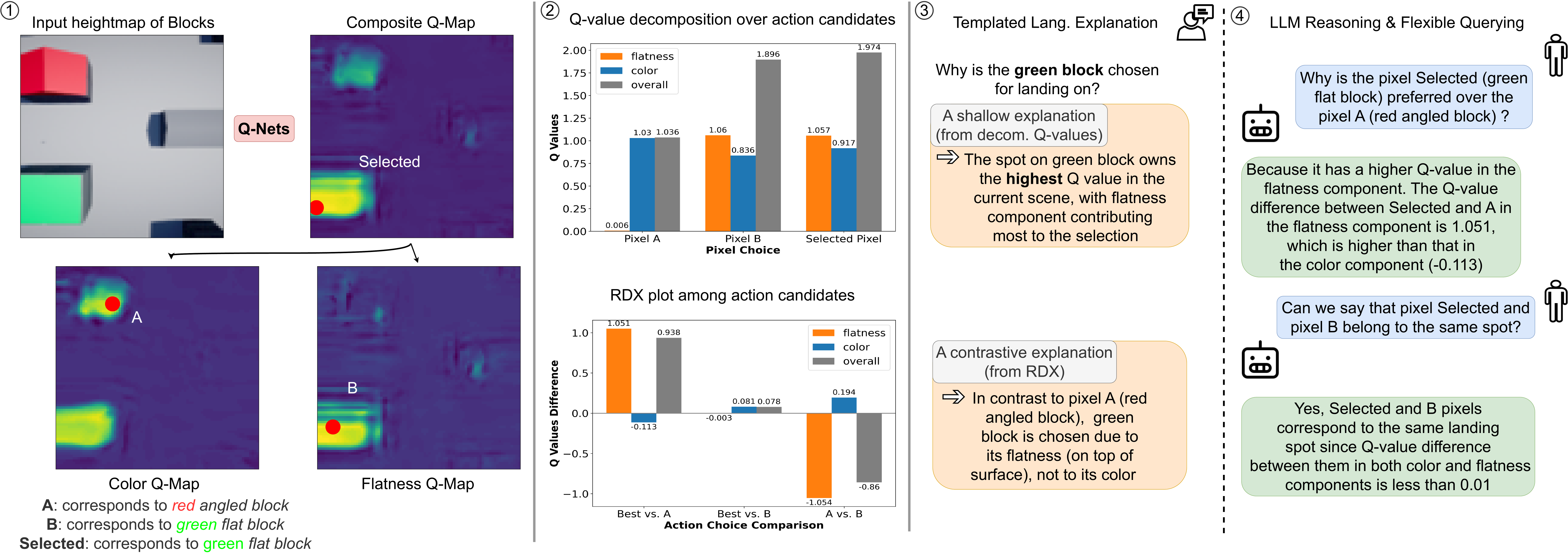}
     \caption{Illustration of the learned Q-Maps for the landing scenario and the multiple modalities of explanations, with the same four parts as in the grasping task.}
    \label{fig:x drone}
\end{figure*}

\textbf{Evaluation Metrics.} 
To evaluate the learned Q-agent's task-solving ability before querying it for an explanation we measure the percentage of correct (highest possible reward) action (pixel) choices out of the total action choices over 10 evaluation runs (i.e., seeds). For the landing, the agent selects one landing spot per run for one scene view, and for the grasping task, it selects pickable spots until 20 objects are grasped.  

\section{Evaluations And Explanations}

\subsection{Inference by Learned Q-Maps}
Test performance of Q-agents in two scenarios is reported in Table \ref{table:success rate of spot inference}, in which we also include the results of a normal Q-learning for comparison, which differs from X-QMap in that the summed sub-rewards are utilized during Q-Map learning. 
The results indicate that both Q-agents have received sufficient training to comprehend the dynamics of the environments and tasks, resulting in consistent performance without sacrificing transparency. Thus, any explanation generated (in either visuals or texts) regarding their behaviour can be deemed reliable to a certain extent.


\begin{table}[th]
\centering
\caption{Success Rate in Two Scenarios over 10 Runs for X-QMap and normal Q-learning respectively.}

\begin{tabular}{c| c c}
\toprule
\thead{Method} & \thead{Landing} & \thead{Grasping}\\[0.5ex] 
\midrule
\midrule
X-QMap   & $90 \pm 0.09$ \% & $90.93 \pm 0.066$ \%\\ 
normal Q-learning  & $90 \pm 0.09$ \% & $87.64 \pm 0.059$ \%\\ 

\bottomrule
\end{tabular}

\label{table:success rate of spot inference}
\end{table}


\subsection{Interpreting Resulting Explanations for Q-Maps} \label{sec: interpreting resulting explanation}

With the resulting Q-Maps in both scenarios, we demonstrate examples of human-generated explanations and LLM-enabled ones.

\textbf{Templated Explanations.} As illustrated in Fig.~\ref{fig:x}, part 1, the final pick falls on the blue cube, among three action candidates (Sec. \ref{sec: visual properties}): \textit{purple bowl}, \textit{green cube}, and \textit{blue cube}. Human experts can elaborate on this choice by analysing the Q-values of components for these actions and visualizing them in bar plots (part 2). The plot of Q-value decomposition shows that the higher activation under the shape component dominates the correct pick (blue cube). From the RDX plots of value differences among action pairs, it is concluded that the object at pixel A is less likely a cube (wrong shape) and the object at pixel B is more likely of inferior colour, though it features the correct shape.

The elaboration above can be easily wrapped into a textual form using templates (Sec. \ref{sec: visual properties}), in which human experts merely need to complete placeholders with object properties, this completion can also be done by an LLM. We argue that this property-guided template is widely applicable to tasks under our Q-Map framework if tasks feature sub-rewards.  
The same mechanism of explanation generation is applied to the landing case, as depicted in Fig.~\ref{fig:x drone}, where explanations are connected to different reward components.

The right parts of Fig.~\ref{fig:x} and Fig.~\ref{fig:x drone} show examples of claims. Unlike the low-level action setting in \cite{cite-key}, our one-step local explanations query about abstract action choices, and thus are not affected by ambiguity issues since Q-agents always have a clear goal signal accompanying their action selection.

\textbf{LLM-generated Language Explanations.}
The interaction between a human user and an LLM model is illustrated in part 4 of Fig. \ref{fig:x} and Fig. \ref{fig:x drone}. The LLM could understand users' questions and offer explanations as a response in its own words. Note the grounding of LLM into both scenarios is done via prompt engineering without the need for fine-tuning (see Table~\ref{tab:prompt}). This showcases how LLM can help convey more versatile explanations to users using its context reasoning capability.  

\subsection{Generalisability and Discussion}
Note other varieties of properties beyond the passive visual perception, such as \textit{impact sound} or \textit{weight}, can also be easily integrated with this framework by regarding them as other reward components of an instructed task. For example, after each execution of the action, the resulting environmental response to the satisfaction of those properties (i.e., binary reward) can be obtained from multi-modal perception modules, e.g., “metallic sound” by an auditory module \cite{Dimiccoli2022RecognizingOS} after striking a metal cube. The independency between the policy learning and perception modules allows other perception modules to be plugged in whenever needed. Further, this framework allows weighing different properties to account for potential preference (e.g., in the landing scenario, a higher weight might be allotted to surface flatness than colour). This can be done by simply scaling the corresponding reward magnitudes.

It may come to another ambiguity in situations where multiple valid object options with identical properties are available for grabbing or landing.
In such cases, the Q-agents being trained may encounter epistemic uncertainty \cite{010c1cec38094a4bb971bc499462763d}, which stems from a lack of additional information. To mitigate this uncertainty, we openly accept the agent's action choice in this work. Nevertheless, our approach can benefit from integrating additional properties (as clarified earlier) from the task environment, aiding in disambiguation and diversifying the rationale behind the action choice.

\section{Conclusions}


The developmental advance of intelligent systems not only brings benefits to individuals through their capabilities but also leads to a growing reliance on them.
In particular, when it comes to increasing trust in decision-making, enhancing the explainability of these systems becomes increasingly crucial.
To address this, we have created the X-QMap framework, which utilizes reward decomposition to untangle the complexity associated with multiple rewarding factors for an agent. The explanation process occurs at a higher semantic level, aligning with human consciousness and interests. This framework incorporates a diverse interface that combines vision and language, ensuring usability and flexibility.

\textbf{Limitations and Future Work.} 
It should be noted that the reliability of the explanations within the Q-Map framework is heavily influenced by the learning capability of the Q-agent. This is not a unique challenge faced only by our proposed framework but is a general concern in the field.
To ensure that the explanations provided by RD are effective in comprehending an agent's behaviour, it is assumed that sub-rewards are implicitly included in the task. Our future work aims to enhance the decomposition of a single task reward by building on existing approaches, such as \cite{lin2020rd}, with the goal of automating the process. 

\addtolength{\textheight}{-2cm}   

\bibliographystyle{./IEEEtran} 
\bibliography{./IEEEabrv,./IEEEexample, XRL}

\end{document}